\PassOptionsToPackage{table, xcdraw}{xcolor}
\documentclass[11pt,a4paper]{article}
\usepackage[hyperref]{conll2018}
\usepackage{times}
\usepackage{latexsym}
\usepackage[utf8]{inputenc}
\usepackage{amsmath}
\usepackage{amsfonts}
\usepackage{amssymb}
\usepackage{pgfgantt}
\usepackage{xcolor}
\usepackage{graphicx}
\usepackage{caption}
\usepackage{subcaption}
\usepackage{csvsimple}
\usepackage[footnotes]{trackchanges}

\usepackage{url}

\aclfinalcopy 

\addeditor{barry} 
\addeditor{steven} 

\title{Using sparse semantic embeddings learned from multimodal text and image data to model human conceptual knowledge}

\author{
	Steven Derby$^1$\ \ \ \ \ \   
	Paul Miller$^1$\ \ \ \ \ \ 
    Brian Murphy$^{1,2}$\ \ \ \ \ \ 
    Barry Devereux$^1$\\\\
  $^1$ Queen's University Belfast, Belfast, United Kingdom \\
  $^2$ BrainWaveBank Ltd., Belfast, United Kingdom \\
  {\{\tt sderby02, \tt p.miller, \tt brian.murphy, \tt b.devereux\}@qub.ac.uk}
}

\date{}

\begin{document}
\maketitle
\begin{abstract}
Distributional models provide a convenient way to model semantics using dense embedding spaces derived from unsupervised learning algorithms. However, the dimensions of dense embedding spaces are not designed to resemble human semantic knowledge. Moreover, embeddings are often built from a single source of information (typically text data), even though neurocognitive research suggests that semantics is deeply linked to both language and perception. In this paper, we combine multimodal information from both text and image-based representations derived from state-of-the-art distributional models to produce sparse, interpretable vectors using \textit{Joint Non-Negative Sparse Embedding}. Through in-depth analyses comparing these sparse models to human-derived behavioural and neuroimaging data, we demonstrate their ability to predict interpretable linguistic descriptions of human ground-truth semantic knowledge.
\end{abstract}







\section{Introduction}

\textit{Distributional Semantic Models} (DSMs) are used to represent semantic information about concepts in a high-dimensional vector space, where each concept is represented as a point in the space such that concepts with more similar meanings are closer together. Unsupervised learning algorithms are regularly employed to produce these models, where learning depends on statistical regularities in the distribution of words, exploiting a theory in linguistics called the \emph{distributional hypothesis}. 
Recent developments in deep learning have resulted in weakly-supervised prediction-based methods, where, for example, a neural network is trained to predict words from surrounding contexts, and the network parameters are interpreted as vectors of the distributional model \cite{mikolov2013distributed}.
Like their counterparts in machine vision, neural network algorithms for DSMs automate feature extraction from highly complex data without prior feature selection 
\cite{krizhevsky2012imagenet, mikolov2013distributed, karpathy2015deep, antol2015vqa}.
Such deep learning techniques have led to state-of-the-art performance in many domains, though this is often at the expense of the interpretability and cognitive plausibility of the learned features \cite{murphy2012learning, zeilerfergus2013}.
Furthermore, these compact, dense embeddings are structurally dissimilar to the way in which humans conceptualise the meanings of words  \cite{mcrae2005semantic}.
One way of drawing interpretability from highly latent data is by transforming it into a sparse representation \cite{faruqui2015sparse, senel2017semantic}. 
Moreover, the design of distributional models has been for the most part unimodal, typically relying on text corpora, even though studies in psychology have shown that human semantic processing is deeply linked with visual perception. 

In cognitive neuroscience, research demonstrates that representations of high-level concepts corresponding to the meanings of nouns and visual objects are widely distributed and overlapping across the cortex \cite{haxby2001distributed, devereux2013representational}, which has opened up research into exploiting machine learning for neurosemantic prediction tasks using distributed semantic models \cite{mitchell2008predicting, huth2016natural, clarke2015predicting, devereux2018}. Such research has helped with both the construction and evaluation of semantic distributional embeddings in computer science \cite{devereux2010using, sogaard2016evaluating}. 
In this paper, we utilise a matrix factorisation algorithm known as \textit{Non-Negative Sparse Embedding} (NNSE) \cite{murphy2012learning}, and an extension known as \textit{Joint Non-Negative Sparse Embedding} (JNNSE) \cite{fyshe2014interpretable} to produce joint sparse multimodal distributions from text and image data. Furthermore, we show that this joint multimodal semantic embedding approach offers a more faithful and parsimonious description of semantics as exhibited in human cognitive knowledge and neurocognitive processing, when compared with dense embeddings learned from the same data.

\section{Related Work}

Much of the research aimed at the sparse decomposition of dense vector spaces is closely associated with the work of \newcite{hoyer2002}, who proposed a \textit{Non-Negative Matrix Factorization technique} (NMF) known as \textit{Non-Negative Sparse Coding} (NNSC) which produces a sparse representation of the original compact matrix.
With the use of new optimisation techniques \cite{mairal2010online}, \newcite{murphy2012learning} later implemented a variation of this approach that forces an L1 penalty on the new sparse matrix, yielding \textit{Non-Negative Sparse Embedding} (NNSE). The purpose of the NNSE algorithm is to generate an embedding that attains the desirable qualities of effectiveness and interpretability (\newcite{murphy2012learning}). Building upon this approach, \newcite{fyshe2014interpretable} extended NNSE to incorporate other sources of semantic information using an extension of NNSE known as \textit{Joint Non-Negative Sparse Embedding} (JNNSE). Their experiments made use of neuroimaging data as an additional source of semantic information, and recent work has seen a push for the incorporation of a broader range of semantic knowledge into DSMs, including semantic knowledge derived from visual image information. 

\newcite{bruni2014multimodal} combined embeddings from text and co-occurrence statistics from data via mining techniques derived from pictures using a procedure known as Visual Bag-of-Words (VBOW). Later this approach was extended by \newcite{kiela2014learning} who incorporated the penultimate layer of modified Convolutional Neural Networks (CNN) to forge a more grounded, semantically faithful model that improved on the state-of-the-art. \newcite{lazaridou2015combining} extend the architecture of the skip-gram model associated with Word2Vec \cite{mikolov2013distributed} to incorporate a measure of visual semantic information by forcing the network to learn linguistic and visual-based features. Instead of performing a context-based prediction task, \newcite{ngiam2011multimodal} combine multimodal information from both audio and image-based information using a stacked autoencoder to reconstruct both modalities with a shared representation layer in the middle of the network. \newcite{silberer2017visually} similarly combine information from multiple modalities from both visual and linguistic data sources by using a stacked autoencoder to reconstruct both types of information separately with a shared representation layer, and a softmax layer connected to the representation layer used to predict the concept characterised by these representations. Rather than trying to construct each modality separately, \newcite{collell2017imagined} make use of a simple perceptron and a neural network to reconstruct the visual modality from pretrained linguistic representations.

Criticism towards traditional distributional models and the benchmarks used to evaluate them \cite{batchkarov2016critique} are now compelling more researchers to consider evaluation techniques that analyse how well distributional models encode different aspects of grounded meaning \cite{lucy2017distributional, collell2016image, gladkova2016intrinsic}.
In particular, one aspect of cognitive plausibility that is lacking in dense representations is in their interpretability, something that could be solved using sparsity \cite{faruqui2015sparse, senel2017semantic}.
In this paper, we combine both text and image-based data in conjunction with matrix factorisation strategies to build sparse and multimodal distributional models, with the goal of demonstrating that these models are more interpretable with respect to human semantic knowledge about concepts. In particular, we show that these models attain a structural composition and semantic representation that is closer to the way humans represent concepts, evaluated using human similarity judgements, human semantic feature knowledge, and neuroimaging data.



\begin{table}[!tb]
\resizebox{\columnwidth}{!}{%
\begin{tabular}{|l|l|r|r|r|}
\hline
\textbf{Modality} & \textbf{Source Embeddings}    & \textbf{\#D}          & \textbf{\#S}     \\ \hline
Text     & GloVe                & 1000         & 200     \\ \hline
Text     & Word2Vec             & 1000         & 200     \\ \hline
Image    & CNN-Mean             & 6144         & 1000    \\ \hline
Image    & CNN-Max              & 6144         & 1000    \\ \hline
Both     & CNN-Mean + GloVe     & 7144         & 200     \\ \hline
Both     & CNN-Max + GloVe      & 7144         & 200     \\ \hline
Both     & CNN-Mean + Word2Vec  & 7144         & 200     \\ \hline
Both     & CNN-Max + Word2Vec   & 7144         & 200     \\ \hline
\end{tabular}
}
\centering
\caption{List of all dense (D) and sparse (S) models used in this paper, and the number of dimensions (\#) in each model.}
\label{modeldims}
\end{table}

\section{Multimodal Representation}
In total, we used sixteen distributional semantic models, eight of which are dense and eight of which are their sparse counterparts. These models are summarized in Table \ref{modeldims}, which describes the eight sources of semantic information (two text-based, two image-based, and four multimodal image+text-based) used to construct both the dense and sparse embedding models. Construction of the eight dense models largely followed \newcite{kiela2014learning}, with eight corresponding sparse models later constructed using JNNSE.  


\subsection{Text-based models}
We implemented two state-of-the-art text-based embedding models, Word2Vec and  GloVe, 
to act as initialisers for our sparse models, following a similar approach to \newcite{faruqui2015sparse}. Both text-based models were trained on 4.5 gigabytes of preprocessed Wikipedia data, with fixed context windows of size 5 and 1000 embedding dimensions. 
The Wikipedia preprocessing was standard and included removal of Wikipedia markup, stop words and non-words, as well as lemmatisation (implemented using standard \emph{NLTK} tools). After model training, the embeddings for each word were normalised to mean zero and unit length, using the L2 norm. Vector normalisation was carried out to ensure magnitudes of the text-based vectors were in line with the image-based vectors, which are normalised by default.

\textbf{GloVe}.  Global Vector for Word Representation \cite{pennington2014glove} is an unsupervised learning algorithm that captures fine-grained semantic information using co-occurrence statistics. It achieves this by constructing real vector embeddings using bilinear logistic regression with non-zero word co-occurrences in the training corpus within a specific context. Our model was trained using a learning rate of $0.05$ over $100$ epochs.

%
%

\textbf{Word2Vec}. Word2Vec \cite{mikolov2013distributed} uses shallow neural networks with negative sampling techniques, which are trained to predict either the word from the context or the context from the word using a fixed window of words as the context. In particular, we choose the CBOW version (predict the word using the context) of this model which was trained using the \emph{gensim} package with the minimum word count threshold set to $0$ (i.e., a vector representation was created for all words in the corpus).

\subsection{Image models}

We make use of the image embeddings constructed by \newcite{kiela2014learning}. In their paper, the AlexNet \cite{krizhevsky2012imagenet} CNN was extended from $1000$ output units to $1512$ outputs, using the additional $512$ object label categories chosen by \newcite{oquab2014learning} and retrained using transfer learning \cite{oquab2014learning}.  This new network was trained using the 2012 version of the \textit{ImageNet Large Scale Visual Recognition Challenge} (ILSVRC) competition dataset with extra images from $512$ other categories, which was then later used to gather embeddings for the ESP game image dataset \cite{von2004labeling}. After training, the network was sliced to remove the final fully-connected softmax layer, in order to retrieve the activation vectors for each image on the penultimate layer. There are systematic differences in the kinds of images that appear in the ImageNet and ESP game training sets. The ImageNet dataset \cite{deng2009imagenet} consists of $12.5$ million images over $22$K different object categories, with each image typically corresponding to a single labelled object (i.e. images do not tend to be cluttered with several objects). In contrast, the ESP game dataset consists of $100000$ images with many labelled objects present in each image.

%
%
To retrieve activation vectors for object categories from the ESP dataset, \newcite{kiela2014learning} used a fair proportional sampling technique: for each object category label, $1000$ images were sampled according to the WordNet \cite{miller1995wordnet} subtree for that concept. If sampling up to $1000$ images was not possible, then the subtree of the concepts hypernym parent node was further sampled until $1000$ images were retrieved.  The activation vector for each of these images was then obtained from the truncated CNN. To retrieve the final embedding vectors for each object label from the sampled activation vectors, \newcite{kiela2014learning} combined the $1000$ activation vectors for each label using two techniques, described below.

\textbf{CNN-Max}. Each word embedding was produced by taking the elementwise maximum value over all 1000 CNN activation vectors obtained for the sampled images with the same label word. 

\textbf{CNN-Mean}. Each word embedding was produced by taking the elementwise average of all 1000 activation vectors associated with the same label word.

All image embeddings are of size $6144$, corresponding to the size of the penultimate layer of the CNN. The embeddings used in our paper correspond to the ESP game labels (which uses a larger number of images, more natural images, and more labels than ImageNet), and all embeddings are normalised to mean zero and L2 unit length before downstream analysis.

\subsection{Multimodal models}
Again following \newcite{kiela2014learning}, we produce four new dense models from combinations of text and image embeddings by simply concatenating the embedding vectors of each model corresponding to each word to create new multimodal text+image embeddings: 
\begin{equation}
	Vec_{multi} = \alpha \times Vec_{text} \hspace{0.1cm} || \hspace{0.1cm} (1 - \alpha) \times Vec_{image}
\end{equation}
\vspace{0.1cm}%
Here, $\alpha$ is a mixing parameter that determines the relative contribution of each modality to the combined semantic space. We set $\alpha = 0.5$, so that text and image sources contribute equally to the combined embeddings.

\section{Sparse matrix factorization}
Following \newcite{faruqui2015sparse}, we use the dense text and image model embeddings as initialisers for corresponding sparse embedding spaces. The embedding vectors are concatenated into an embedding matrix for each model, with the number of rows corresponding to the number of words in their respective lexicons, and the number of columns corresponding to the embedding dimensionality. 

To produce the new sparse representations, we use the NNSE matrix factorisation technique
\footnote{ Non-Negative Sparse Embedding code was kindly provided by Partha Talukdar.}
(\newcite{murphy2012learning}) which maps a dense word-feature matrix $X$ to a non-negative sparse matrix $A$ with an identical lexicon. 
Let $X \in \mathbb{R}^{w \times k}$ be an embedding matrix, where $w$ is the number of words, and $k$ is the embedding dimension size. 
NNSE factorises $X$ into two matrices, a dictionary transformation matrix $D \in \mathbb{R}^{p \times k}$ and the sparse matrix $A \in \mathbb{R}^{w \times p}$ by minimising the objective function:
\begin{equation}
	arg\min_{D ,A} \sum_{i=1}^{w}||X_{i,:} - A_{i,:} \times D||^{2} + \lambda ||A_{i,:}||_{1}
\end{equation}
subject to the constraints
\begin{align*}
	 \quad D_{i,:} D_{i,:}^{T} \leq 1, \, \forall \, 1 \leq i \leq p \\
	 A_{i,j} \geq 0, \, \forall \, 1 \leq i \leq w, \, \forall \, 1 \leq j \leq p
\end{align*}
which ensure sparse and non-trivial solutions for $A$ (\newcite{murphy2012learning}).

NNSE has been extended as a method to combine multiple dense word-feature matrices $X \in \mathbb{R}^{w_x \times k}$ and $Y \in \mathbb{R}^{w_y \times n}$ into a single non-negative sparse matix, an extension called Joint Non-Negative Sparse Embedding (JNNSE; \newcite{fyshe2014interpretable}). 
Although JNNSE can be used with feature matrices with different lexicons, in this paper we take only the $w$ rows of the two matrices that correspond to the intersection of words used to build the two embedding models and a set of $2234$ unique concept words taken from the four similarity evaluation datasets discussed in the next section. 
JNNSE gives a new joint sparse feature matrix $A \in \mathbb{R}^{w \times p}$ by minimising the objective function:%
\begin{equation}
\begin{aligned}
	arg\min_{D^{(x)}, D^{(y)} , A} \sum_{i=1}^{w}||X_{i,:} - A_{i,:} \times D^{(x)}||^{2} \\ 
		+ \sum_{i=1}^{w}||Y_{i,:} - A_{i,:} \times D^{(y)}||^{2} + \lambda ||A_{i,:}||_{1}
\end{aligned}
\end{equation}where%
\begin{align*}
	\quad D_{i,:}^{(x)}  D_{i,:}^{(x)^{T}} \leq 1, \, \forall \, 1 \leq i \leq p \\
		  \quad D_{i,:}^{(y)} D_{i,:}^{(y)^{T}} \leq 1, \, \forall \, 1 \leq i \leq p \\
	 A_{i,j} \geq 0, \, \forall \,1 \leq i \leq w, \, \forall \, 1 \leq j \leq p
\end{align*}
\vspace{-0.3cm}%

\begin{figure*}[!tb]
	\begin{subfigure}{.5\textwidth}
		\vspace{-1cm}
		\includegraphics[height = 6.0cm, width = 8cm]{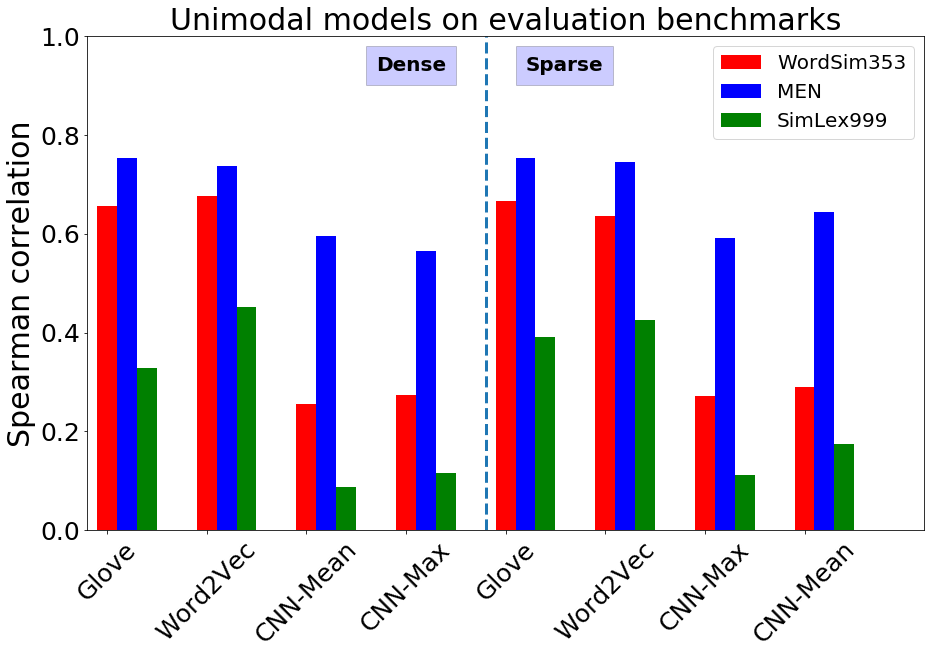}
	\end{subfigure}
	\begin{subfigure}{.5\textwidth}
		\includegraphics[height = 7cm, width = 8cm]{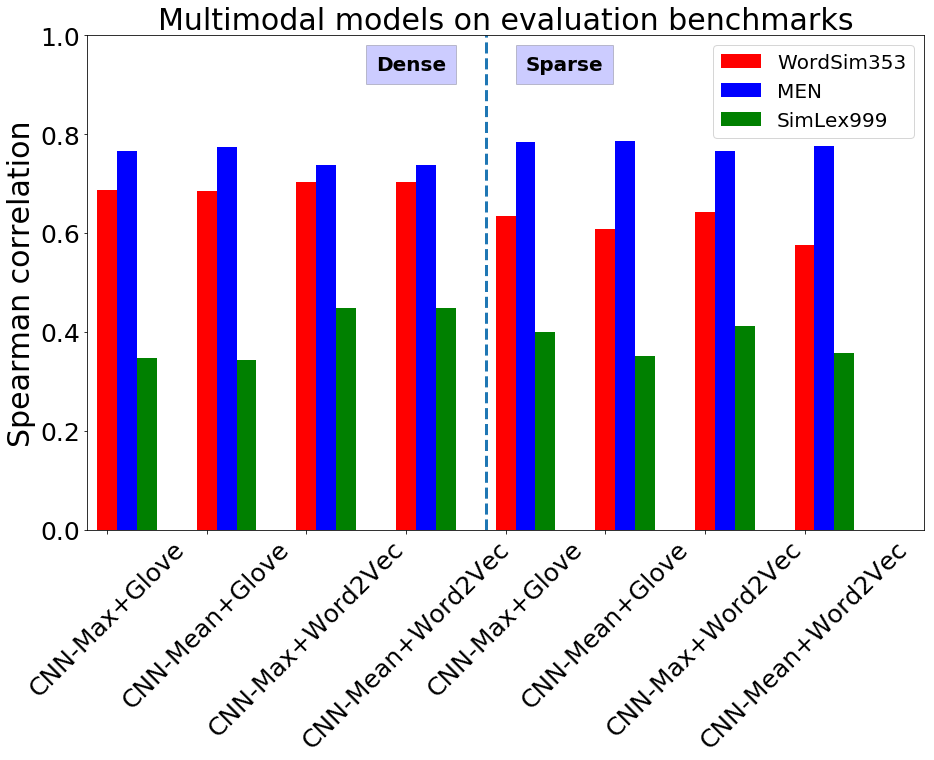}
	\end{subfigure}
	\caption{Results for the dense and sparse embeddings for the three semantic similarity benchmarks, for the eight unimodal models (left panel) and the eight multimodal image+text models (right panel).}
	\label{semsim}
\end{figure*}

For the NNSE factorization of each of the four initial dense unimodal text and image models (GloVe, Word2Vec, CNN-Mean and CNN-Max), the sparsity parameter $\lambda$ was set to 0.05 and each model's dimensionality ($p$) was reduced down from its original size by a factor of approximately 5; the text embedding size was reduced to $200$ and both image model embedding sizes were reduced to $1000$ (see Table \ref{modeldims}). 

To create sparse multimodal models corresponding to the concatenated multimodal dense models, four new models were produced using Equation 3. These models were constructed by combining all combinations of pruned image and text-based models through JNNSE to produce sparse embeddings of size $200$ from their original dimensions of $6144$ and $1000$ respectively. The sparsity parameter $\lambda$ was set to $0.025$. 
Though all sparse embedding matrices are calculated over a smaller lexicon and have a much smaller embedding size compared to the original dense embeddings, in the next section, we investigate how these models still produce competitive results on semantic evaluation benchmarks, including neurocognitive data.

\begin{table*}[!tb]
\resizebox{\textwidth}{!}{%
\begin{tabular}{|l|p{1.7cm}|l|p{1.8cm}|l|l|l|}
\hline
\textbf{Model}                        & \textbf{En\-cyc\-lo\-ped\-ic}             & \textbf{Functional}                 & \textbf{Taxonomic}            & \textbf{Visual}                  & \textbf{Other Perceptual}          & \textbf{Overall}                    \\ \hline
\textbf{CNN-Mean}                  & 23.479                         & 28.309                         & 45.756                         & 31.256                         & 26.467                         & 29.244                         \\ \hline
\textbf{CNN-Max}                   & 22.878                         & 28.765                         & 50.140                         & 32.843                         & 27.508                         & 30.202                         \\ \hline
\textbf{GloVe}                     & 30.870                         & 37.176                         & 61.517                         & 35.909                         & \cellcolor[HTML]{38FFF8}38.385 & 36.984                         \\ \hline
\textbf{Word2Vec}                  & 27.494                         & 30.372                         & 55.455                         & 32.298                         & 32.800                         & 32.363                         \\ \hline
\textbf{GloVe NNSE}                & 31.171                         & 34.645                         & 59.497                         & 35.066                         & 36.738                         & 35.880                         \\ \hline
\textbf{Word2Vec NNSE}             & 29.662                         & 34.320                         & 55.073                         & 35.302                         & 33.261                         & 34.956                         \\ \hline
\textbf{CNN-Max NNSE}              & 15.320                         & 17.138                         & 26.263                         & 19.646                         & 17.453                         & 18.279                         \\ \hline
\textbf{CNN-Mean NNSE}             & 15.996                         & 18.297                         & 27.330                         & 20.954                         & 18.376                         & 19.339                         \\ \hline
\textbf{CNN-Max + GloVe}           & 30.669                         & 37.404                         & 63.887                         & 35.790                         & 36.077                         & 36.760                         \\ \hline
\textbf{CNN-Mean + GloVe}          & 31.560                         & 38.441                         & 64.459                         & 36.675                         & 36.625                         & 37.637                         \\ \hline
\textbf{CNN-Max + Word2Vec}        & 22.114                         & 24.653                         & 51.471                         & 27.566                         & 27.332                         & 27.088                         \\ \hline
\textbf{CNN-Mean + Word2Vec}       & 22.057                         & 24.780                         & 51.926                         & 27.527                         & 27.407                         & 27.124                         \\ \hline
\textbf{CNN-Max + GloVe JNNSE}     & 32.481                         & \cellcolor[HTML]{38FFF8}38.787 & 63.669                         & 39.848                         & 36.245                         & \cellcolor[HTML]{38FFF8}39.080 \\ \hline
\textbf{CNN-Mean + GloVe JNNSE}    & 31.104                         & 38.009                         & \cellcolor[HTML]{38FFF8}64.866 & \cellcolor[HTML]{38FFF8}40.267 & 35.998                         & 38.784                         \\ \hline
\textbf{CNN-Max + Word2Vec JNNSE}  & \cellcolor[HTML]{38FFF8}32.718 & 38.601                         & 61.493                         & 39.663                         & 36.496                         & 38.901                         \\ \hline
\textbf{CNN-Mean + Word2Vec JNNSE} & 31.084                         & 36.939                         & 57.659                         & 38.145                         & 33.436                         & 37.057                         \\ \hline
                        
\end{tabular}
}
\centering
\caption{Average cross-validation F1 $\times 100$ scores for each model. The blue highlighting indicates the model that scores the highest on each property class.}
\label{f1scores}
\end{table*}

\section{Experiments}
\vspace{-0.1cm}
The aim of our experiments is to compare the quality of the dense and sparse unimodal and multimodal embedding models, with a focus on their ability to explain human-derived semantic data. We use several qualitatively different analyses of how well the models explain human-derived measures of semantic representation and processing. In the results that follow, we first demonstrate that sparse multimodal models are competitive with larger dense embedding models on standard semantic similarity evaluation benchmarks. We then investigate whether the underlying representations of the sparse, multimodal models are more easily interpreted in terms of human semantic property knowledge about familiar concepts, by evaluating the models' ability to predict predicates describing property knowledge found in human property norm data. Finally, we evaluate the models' ability to predict human brain activation data.  

\subsection{Semantic similarity benchmarks}
A widely used evaluation technique for distributional models is the comparison with human semantic similarity rating benchmarks. We evaluate our models on three popular datasets which each reflect slightly different intuitions about semantic similarity. 

\textbf{WordSim353} \cite{finkelstein2001placing} consists of $353$ word pairs with human ratings indicating how related the two concepts in each pair are. The definition of similarity is left quite ambiguous for the human annotators, and words which share any kind of association tend to receive high scores. 

\textbf{MEN} \cite{bruni2012distributional} consists of $3000$ word pairs with human ratings of how semantically related each pair of concepts are. Pairs with high scores tend to be linked more by semantic relatedness than by similarity; for example, the words ``coffee'' and ``cup'' are semantically related (even though a cup is not similar to coffee). Semantic relatedness often corresponds to meronym or holonym concept pairings (e.g. ``finger'' - ``hand''). 

\textbf{SimLex999}. \cite{hill2015simlex} is a comprehensive and modern benchmark consisting of $999$ pairs of words with human ratings of semantic similarity. Semantic similarity tends to reflect words with shared hypernym relations between concept pairs (e.g. ``coffee'' \& ``tea'' are more similar than ``coffee'' \& ``cup''). 

In evaluating against the benchmarks, we use the intersection of the words occurring in the benchmarks and the words used in creating our embeddings. Not all words used in the similarity benchmarks appear in our word embedding models, although the overlap is quite high\footnote{Atleast $83\%$ for SimLex999, $81\%$ WordSim353 and $94\%$ for MEN.}. 
Evaluations in the next section are based on the subsets of word-pairs for which we have embedding vectors for each word.

\subsection{Semantic Similarity Results}

Figure \ref{semsim} shows the results for all $16$ models on the three evaluation datasets. Even with their significant dimensionality reduction and forced sparsity regularisation, the sparse (NNSE) unimodal text and image-based models perform comparatively with their original dense counterparts, with better results for the sparse unimodal models on several of the benchmarks. The JNNSE models perform comparably to their dense counterparts, with performance on MEN slightly improved, performance on WordSem353 marginally worse, and performance on SimLex999 approximately the same (in spite of the JNNSE models having less than $1/35$ times the number dimensions of their sparse counterparts)\footnote{In order to ensure that the dense models were not disadvantaged by having more dimensions, we also trained dense text models with $200$ dimensions and found that these did not perform better than the $1000$-dimensional models. Furthermore, we applied SVD to each of the $1000$-dimensional dense models to reduce the number of dimensions to $200$ but again found the results to be worse than the results for both the $1000$-dimensional dense models and the sparse models.}.  Finally, the combined text+image multimodal embeddings are better than unimodal models overall at fitting the similarity rating data. The results on these conventional benchmarks suggest redundancy in the dense embedding representations, with the sparse embeddings providing a parsimonious representation of semantics that retains information about semantic similarity. Moreover, multimodal models combining both linguistic and perceptual experience better account for human similarity judgements.  



\subsection{Property norm prediction}

Following \newcite{collell2016image} and \newcite{lucy2017distributional}, we make use of a dataset of human-derived property norms for a set of concepts and analyse how well our distributional models can predict human-elicited property knowledge for words. We use the CSLB property norms (\newcite{devereux2014centre}), a dataset of semantic features for a set of $541$ noun concepts, elicited by participants in a large-scale property norming study. (For example, for ``apple", properties include \emph{is-a-fruit}, \emph{is-red}, \emph{grows-on-trees}, \emph{has-seeds}, \emph{is-round}, etc.). For each embedding model, we train an $L2$ regularised logistic regression classifier for each property that predicts whether the property is true for a given concept. 

The human-elicited property$\times$concept matrix is sparse; most properties are not true of most concepts. For the logistic regression model trained for each semantic property, we therefore balance positive and negative training items by weighting coefficients inversely proportional to the frequency of the two classes. Properties which are true of less than five concepts (across the set of concepts appearing in both the CSLB norms and our embedding models) were removed, to ensure sufficient positive and negative training cases across concepts. To evaluate the logistic regression models' ability to predict human property knowledge for held-out concepts, we used $5$-fold cross-validation with stratified sampling to ensure that at least one positive case occurred in each test set. Using the embedding dimensions as training data, we train on the $4$ folds and test on the final fold, and evaluate the logistic regression classifier by taking the average F1 score over all the test folds. For subsequent analysis of the fitted regression models for each property, the semantic properties were partitioned into the five general classes given in \newcite{devereux2014centre}. These property classes were \emph{visual} (e.g. \emph{is-green}; \emph{is-round}), \emph{functional} (e.g. \emph{is-eaten}; \emph{used-for-cutting}), \emph{taxonomic} (e.g. \emph{is-a-fruit}; \emph{is-a-tool}), encyclopedic (\emph{has-vitimans}; \emph{uses-fuel}), and \emph{other-perceptual} (e.g. \emph{is-tasty}; \emph{is-loud}). We hypothesised that properties of different types would differ in how accurately they could be predicted from the different embedding models, given the different sources of information available in the models (for example, visual properties may be more predictable from models trained with image data; see also \newcite{collell2016image}).

Table \ref{f1scores} shows the average F1 scores overall properties and over each of the five property categories. Since the dense and sparse models trained on the same source data (text, images, or text+images) encode similar information, they perform similarly on the task of predicting human semantic property knowledge. However, sparse multimodal models (the last four rows of the table) are the top scoring models for four of the five property categories, and over the full set of properties (last column of Table \ref{f1scores}) the top three models are all sparse and multimodal. These results indicate that sparse multimodal embeddings are superior to their single modality and dense counterparts in their ability to predict interpretable semantic properties corresponding to elements of human conceptual knowledge.

\subsection{Interpretating embedding dimensions in terms of semantic properties}
\begin{figure}[!tb]
	\centering
	\includegraphics[height = 6cm , width = 8cm]{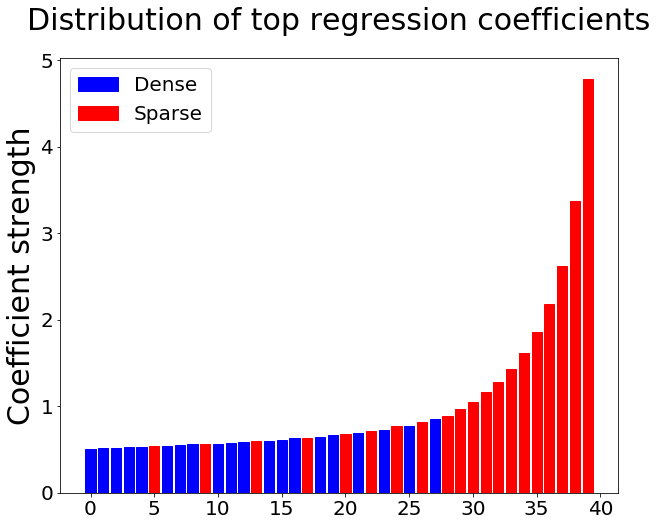}
	\caption{The ranking of the top $20$ model coefficients for the logistic regression classifiers trained on each feature, for the dense GloVe + CNN-Mean model (blue bars), and the joint sparse GloVe + CNN-Mean model (red bars).
}
\label{regcoeffs}
\end{figure}

Information about a specific semantic property can be stored latently over the dimensions of a semantic embedding model, such that the semantic property can be reliably decoded given an embedding vector, as tested in the previous section. However, a stronger test of how closely an embedding model relates to human-elicited conceptual knowledge is to investigate whether the embedding dimensions encode interpretable, human-like semantic properties directly. In other words, does an embedding model learn a set of basis vectors for the semantic space that corresponds to verbalisable, human semantic properties like \emph{is-round}, \emph{is-a-fruit}, and so on?    
To address this question, we evaluated how the dense and sparse embeddings differ in their degree of correspondence to the property norms by analysing the fitted parameters of our property prediction logistic regression classifiers. For each embedding model and semantic property, we average the fitted parameters in the logistic regression models across cross-validation iterations and extract the $20$ parameters with the highest average magnitude. For each property, we store these $20$ parameters in a vector sorted by decreasing magnitude. If a particular semantic property is decodable directly from only one (or very few) embedding dimensions, then the magnitude of the first element (or few elements) of the sorted parameter vector will be very high. Over all properties, we then apply element-wise averaging of the sorted parameter vectors.
Figure \ref{regcoeffs} shows the magnitudes of these 20 averaged parameters for the dense and sparse multimodal GloVe+CNN-Mean models\footnote{The results are similar for all other pairs of sparse and dense models.}. As we can see, the dense model has a more uniform distribution, indicating that the information is highly diffuse over the dimensions of the dense embedding space. Conversely, the top few parameters for the sparse model have very high magnitude, indicating that, on average, information about semantic properties tend to be strongly associated with a small number of dimensions in the sparse space.

\begin{table*}[!tb]
\resizebox{\textwidth}{!}{%
\begin{tabular}{|l|l|l|p{2cm}|p{2cm}|p{2cm}|p{2cm}|p{2cm}|p{2cm}|}
\hline
              & \textbf{GloVe} & \textbf{Word2Vec} & \textbf{CNN-Max} & \textbf{CNN-Mean} & \textbf{CNN-Max + GloVe} & \textbf{CNN-Mean + GloVe} & \textbf{CNN-Max + Word2Vec} & \textbf{CNN-Mean + Word2Vec} \\ \hline
\textbf{fMRI (S)} & 0.654       & 0.652          & 0.641     & 0.647         & 0.662              & \textbf{0.686}              & 0.649                 & 0.671                 \\ \hline
\textbf{fMRI (D)} & 0.670       & 0.676          & 0.654      & 0.651       & 0.673              & 0.677               & 0.676                 & 0.676                  \\ \hline
\textbf{MEG (S)}  & 0.664       & 0.669          & 0.651      & 0.641       & 0.671              & 0.668              & 0.675                 & 0.665                  \\ \hline
\textbf{MEG (D)}  & 0.680       & 0.664           & 0.654      & 0.643       & \textbf{0.684}              & \textbf{0.684}               & 0.664                 & 0.664                  \\ \hline
\end{tabular}
}
\caption{Results of all sparse (S) and dense (D) models on $2$ vs. $2$ tests against the fMRI and MEG neuroimaging data, averaged over participants.}
\label{2v2table}
\end{table*}

As a further test of how well dimensions of embedding models correspond to human semantic knowledge, we calculated pairwise correlations, across concepts, between embedding dimensions and properties. For a given semantic property, we can test which of two embedding models best encode that semantic property in a single dimension -- an embedding model that more directly matches the property norm data will tend to have a dimension that correlates more strongly with that property than \emph{any} dimension of a model that encodes information about that property more latently. For this analysis, we first filtered the set of concepts in the dense models to include only the concepts in the CSLB norms, and recalculated the (J)NNSE sparse models over these concepts only. We tuned the sparsity parameter so that the sparsity of the sparse embedding models closely matched the sparsity of CSLB concept-property matrix (97\% sparse), and kept the dimensionality of the sparse embeddings the same as our original sparse models.
Let $v_{P}$ be the values for a property $P$ for each concept in the CSLB norms, and let $M_{d}$ and $M_{s}$ represent the set of embedding columns for a dense model and its sparse counterpart respectively. Then for each property $P$, we evaluate the inequality
\[
	max_{c \in M_{d}}(\rho(c, v_{P})) < max_{c \in M_{s}}(\rho(c, v_{P})) 
\]
where $\rho$ is the Spearman correlation. We count the proportion of times the inequality is true across all properties in the norms, repeat this for each of the eight dense models and their sparse counterparts, and calculate the average. The results show that the sparse models have the most correlated dimension $63.2\%$ of the time.
In order to ensure that the dense models were not disadvantaged by having more dimensions (and to test that the sparsity constraint rather than dimensionality reduction was the reason for the superior performance of the sparse models), we used SVD on all dense models to reduce the dimensions down to the same size as their sparse counterparts and reran the test. Here the results show that the sparse models have the most correlated dimension $81.1\%$ of the time, indicating that the sparse models do learn semantics-encoding dimensions from the dense models that more closely correspond to human-derived property knowledge.   

\begin{table*}[!tb]
\resizebox{\textwidth}{!}{%
\begin{tabular}{|l|l|l|p{2cm}|p{2cm}|p{2cm}|p{2cm}|p{2cm}|p{2cm}|}
\hline
              & \textbf{GloVe} & \textbf{Word2Vec} & \textbf{CNN-Max} & \textbf{CNN-Mean} & \textbf{CNN-Max + GloVe} & \textbf{CNN-Mean + GloVe} & \textbf{CNN-Max + Word2Vec} & \textbf{CNN-Mean + Word2Vec} \\ \hline
\textbf{fMRI (D)} & 0.162       & 0.164          & 0.145      & 0.151       & 0.150              & 0.152               & 0.152                 & 0.155                  \\ \hline
\textbf{fMRI (S)} & 0.138       & 0.136          & 0.140      & 0.144       & 0.139              & 0.140               & 0.154                 & \textbf{0.168}                  \\ \hline
\textbf{MEG (D)}  & 0.163       & 0.161          & 0.163      & 0.158       & 0.162              & 0.158               & \textbf{0.168}                 & 0.162                  \\ \hline
\textbf{MEG (S)}  & \textbf{0.168}       & 0.152          & 0.149       & 0.149       & 0.152              & 0.157               & 0.145                 & 0.147                  \\ \hline
\end{tabular}
}
\caption{Average RSA results (Spearman's $\rho$) for all sparse (S) and dense (D) models.}
\label{rsatable}
\end{table*}

\subsection{Evaluation on brain data}
For our final set of analysis, we tested how closely each of the eight dense and eight sparse models relate to neurocognitive processing in the human brain. We used BrainBench \cite{xu2016brainbench}, an evaluation benchmark for semantic models that allows us to evaluate each model's ability to predict patterns of activation in neuroimaging data. The BrainBench dataset includes brain activation data recorded using two complementary neuroimaging modalities: \textbf{fMRI} (which measures cerebral blood oxygenation and which has relatively good spatial resolution but poor temporal resolution) and \textbf{MEG} (which measures aggregate magnetic field changes induced by neural activity and which has good temporal resolution but poorer spatial resolution). The neuroimaging data in both modalities are taken from nine participants that viewed pictures of $60$ different concepts. 

The first step is to transform the embedding matrices and the brain activation data into a format that more readily facilitates comparison of these two very different kinds of data. 
For each distributional model, we calculated the pairwise correlation between concepts to produce the $60\times60$ similarity matrix $M$ where each element $M_{i,j}$ 
in the matrix is the correlation between the embedding vectors of the distributional model for the $i$-th and $j$-th concepts. In Brainbench, the brain data is already preprocessed and transformed into such a representation for both the fMRI and MEG neuroimaging modalities, giving a $60\times60$ similarity matrix for each participant for both modalities. The next step for BrainBench evaluation is to perform a ``$2$ vs.~$2$'' test between each distributional model and the brain data. Let $M_{D}$ and $M_{B}$ be the similarity matrices associated with a distributional semantic model and a participant's brain data respectively. Let $r$ be the Pearson correlation function, then a $2$ vs. $2$ test is a positive case for any two pairs of concepts $w_{1}$ and $w_{2}$ if
\begin{align*}
	r(M_{D}(w_{1}), M_{B}(w_{1})) + r(M_{D}(w_{2}), M_{B}(w_{2})) \\
	> r(M_{D}(w_{1}),M_{B}(w_{2})) + r(M_{D}(w_{2}), M_{B}(w_{1}))
\end{align*}
where $M_{D}(w_{1})$ and $M_{D}(w_{2})$ denote the rows of values corresponding to the concepts $w_{1}$ and $w_{2}$ respectively, omitting the columns that correspond to the correlation between $w_{1}$ and $w_{2}$. This $2$ vs.~$2$ test is performed on all pairs of the $60$ concepts, to obtain the proportion of positive cases for the pair $M_{D}$ and $M_{B}$. The distributional models are evaluated against all brain-based representations and averaged by imaging modality. The results for both sparse and dense models are displayed in Table \ref{2v2table}. For the fMRI data, the model with the highest average $2$ vs.~$2$ test score is the sparse multimodal GloVe+CNN-Max embedding, whilst on the MEG data the highest scoring model is a tie between the dense multimodal GloVe+CNN-Max embedding and the dense multimodal GloVe+CNN-Mean embedding. The results demonstrate that semantic distributional models that encode a range of different information are better at making statistically significant predictions on brain data. In general, the multimodal models do better than the unimodal text and image models at fitting the brain data.

Finally, we computed the direct correlation between the representations $M_{D}$ and $M_{B}$, using the technique of \textit{Representational Semantic Analsysis} (RSA) \cite{kriegeskorte2008representational} commonly employed in cognitive neuroscience. Given that $M_{D}$ and $M_{B}$ have the same number of words and word indexing (words associated with certain rows and columns are shared across representations), we take the Spearman's correlation between the flattened upper triangular similarity matrices of these two representations for each pair of DSM and brain dataset\footnote{Usually RSA is performed on a new matrix produced by subtracting an $N \times N$ matrix of all $1$'s from these concept matrices $M_{D}$ and $M_{B}$, where $N$ is the number of shared concepts. Such a representation is known as a \textit{Representational Dissimilarity Matrix} (RDM), although here we follow \newcite{xu2016brainbench} and use similarities.}. 
%

For a given distributional model, we average all Spearman correlation values across the nine participants for each imaging modality; the results are presented in Table \ref{rsatable}. 
The results show that sparse models give the closest representation to both fMRI and MEG data, with the multimodal sparse word2vec+CNN-Mean model best fitting the fMRI data, and the sparse GloVe model best fitting the MEG data. These results indicate that semantic model sparsity is an important property reflected in neurocognitive semantic representations.

\section{Conclusion}

In this paper, we have demonstrated the representational potential of sparse multimodal distributional models using several qualitatively different and complimentary evaluation tasks that are derived from human data: semantic similarity ratings, conceptual property knowledge, and neuroimaging data. We show that both sparse and multimodal embeddings achieve a more faithful representation of human semantics than dense models constructed from a single information source. 

\section{Acknowledgements}
We would like to thank Partha Talukdar for generously providing us with the code for the Non-Negative Sparse Embedding algorithm. We would also like to thank Alona Fyshe for providing the Joint Non-Negative Sparse Embedding code.

\bibliographystyle{acl_natbib_nourl}
\bibliography{Bibliography}

\begin{thebibliography}{41}
\expandafter\ifx\csname natexlab\endcsname\relax\def\natexlab#1{#1}\fi

\bibitem[{Antol et~al.(2015)Antol, Agrawal, Lu, Mitchell, Batra,
  Lawrence~Zitnick, and Parikh}]{antol2015vqa}
Stanislaw Antol, Aishwarya Agrawal, Jiasen Lu, Margaret Mitchell, Dhruv Batra,
  C~Lawrence~Zitnick, and Devi Parikh. 2015.
\newblock Vqa: Visual question answering.
\newblock In \emph{Proceedings of the IEEE International Conference on Computer
  Vision}, pages 2425--2433.

\bibitem[{Batchkarov et~al.(2016)Batchkarov, Kober, Reffin, Weeds, and
  Weir}]{batchkarov2016critique}
Miroslav Batchkarov, Thomas Kober, Jeremy Reffin, Julie Weeds, and David Weir.
  2016.
\newblock A critique of word similarity as a method for evaluating
  distributional semantic models.

\bibitem[{Bruni et~al.(2012)Bruni, Boleda, Baroni, and
  Tran}]{bruni2012distributional}
Elia Bruni, Gemma Boleda, Marco Baroni, and Nam-Khanh Tran. 2012.
\newblock Distributional semantics in technicolor.
\newblock In \emph{Proceedings of the 50th Annual Meeting of the Association
  for Computational Linguistics: Long Papers-Volume 1}, pages 136--145.
  Association for Computational Linguistics.

\bibitem[{Bruni et~al.(2014)Bruni, Tran, and Baroni}]{bruni2014multimodal}
Elia Bruni, Nam-Khanh Tran, and Marco Baroni. 2014.
\newblock Multimodal distributional semantics.
\newblock \emph{Journal of Artificial Intelligence Research}, 49:1--47.

\bibitem[{Clarke et~al.(2015)Clarke, Devereux, Randall, and
  Tyler}]{clarke2015predicting}
Alex Clarke, Barry~J. Devereux, Billi Randall, and Lorraine~K. Tyler. 2015.
\newblock Predicting the time course of individual objects with meg.
\newblock \emph{Cerebral Cortex}, 25(10):3602--3612.

\bibitem[{Collell and Moens(2016)}]{collell2016image}
Guillem Collell and Marie-Francine Moens. 2016.
\newblock Is an image worth more than a thousand words? on the fine-grain
  semantic differences between visual and linguistic representations.
\newblock In \emph{Proceedings of COLING 2016, the 26th International
  Conference on Computational Linguistics: Technical Papers}, pages 2807--2817.
  The COLING 2016 Organizing Committee.

\bibitem[{Collell et~al.(2017)Collell, Zhang, and Moens}]{collell2017imagined}
Guillem Collell, Ted Zhang, and Marie-Francine Moens. 2017.
\newblock Imagined visual representations as multimodal embeddings.
\newblock In \emph{AAAI}, pages 4378--4384.

\bibitem[{Deng et~al.(2009)Deng, Dong, Socher, Li, Li, and
  Fei-Fei}]{deng2009imagenet}
Jia Deng, Wei Dong, Richard Socher, Li-Jia Li, Kai Li, and Li~Fei-Fei. 2009.
\newblock Imagenet: A large-scale hierarchical image database.
\newblock In \emph{Computer Vision and Pattern Recognition, 2009. CVPR 2009.
  IEEE Conference on}, pages 248--255. IEEE.

\bibitem[{Devereux et~al.(2010)Devereux, Kelly, and
  Korhonen}]{devereux2010using}
Barry Devereux, Colin Kelly, and Anna Korhonen. 2010.
\newblock Using fmri activation to conceptual stimuli to evaluate methods for
  extracting conceptual representations from corpora.
\newblock In \emph{Proceedings of the NAACL HLT 2010 First Workshop on
  Computational Neurolinguistics}, pages 70--78. Association for Computational
  Linguistics.

\bibitem[{Devereux et~al.(2013)Devereux, Clarke, Marouchos, and
  Tyler}]{devereux2013representational}
Barry~J Devereux, Alex Clarke, Andreas Marouchos, and Lorraine~K Tyler. 2013.
\newblock Representational similarity analysis reveals commonalities and
  differences in the semantic processing of words and objects.
\newblock \emph{Journal of Neuroscience}, 33(48):18906--18916.

\bibitem[{Devereux et~al.(2018)Devereux, Clarke, and Tyler}]{devereux2018}
Barry~J Devereux, Alex Clarke, and Lorraine~K Tyler. 2018.
\newblock Integrated deep visual and semantic attractor neural networks predict
  fmri pattern-information along the ventral object processing pathway.
\newblock \emph{Scientific Reports}, 8:10636.

\bibitem[{Devereux et~al.(2014)Devereux, Tyler, Geertzen, and
  Randall}]{devereux2014centre}
Barry~J Devereux, Lorraine~K Tyler, Jeroen Geertzen, and Billi Randall. 2014.
\newblock The centre for speech, language and the brain (cslb) concept property
  norms.
\newblock \emph{Behavior research methods}, 46(4):1119--1127.

\bibitem[{Faruqui et~al.(2015)Faruqui, Tsvetkov, Yogatama, Dyer, and
  Smith}]{faruqui2015sparse}
Manaal Faruqui, Yulia Tsvetkov, Dani Yogatama, Chris Dyer, and Noah Smith.
  2015.
\newblock Sparse overcomplete word vector representations.
\newblock \emph{arXiv preprint arXiv:1506.02004}.

\bibitem[{Finkelstein et~al.(2001)Finkelstein, Gabrilovich, Matias, Rivlin,
  Solan, Wolfman, and Ruppin}]{finkelstein2001placing}
Lev Finkelstein, Evgeniy Gabrilovich, Yossi Matias, Ehud Rivlin, Zach Solan,
  Gadi Wolfman, and Eytan Ruppin. 2001.
\newblock Placing search in context: The concept revisited.
\newblock In \emph{Proceedings of the 10th international conference on World
  Wide Web}, pages 406--414. ACM.

\bibitem[{Fyshe et~al.(2014)Fyshe, Talukdar, Murphy, and
  Mitchell}]{fyshe2014interpretable}
Alona Fyshe, Partha~P Talukdar, Brian Murphy, and Tom~M Mitchell. 2014.
\newblock Interpretable semantic vectors from a joint model of brain-and
  text-based meaning.
\newblock In \emph{Proceedings of the conference. Association for Computational
  Linguistics. Meeting}, volume 2014, page 489. NIH Public Access.

\bibitem[{Gladkova and Drozd(2016)}]{gladkova2016intrinsic}
Anna Gladkova and Aleksandr Drozd. 2016.
\newblock Intrinsic evaluations of word embeddings: What can we do better?
\newblock In \emph{RepEval@ACL}.

\bibitem[{Haxby et~al.(2001)Haxby, Gobbini, Furey, Ishai, Schouten, and
  Pietrini}]{haxby2001distributed}
James~V Haxby, M~Ida Gobbini, Maura~L Furey, Alumit Ishai, Jennifer~L Schouten,
  and Pietro Pietrini. 2001.
\newblock Distributed and overlapping representations of faces and objects in
  ventral temporal cortex.
\newblock \emph{Science}, 293(5539):2425--2430.

\bibitem[{Hill et~al.(2015)Hill, Reichart, and Korhonen}]{hill2015simlex}
Felix Hill, Roi Reichart, and Anna Korhonen. 2015.
\newblock Simlex-999: Evaluating semantic models with (genuine) similarity
  estimation.
\newblock \emph{Computational Linguistics}, 41(4):665--695.

\bibitem[{Hoyer(2002)}]{hoyer2002}
Patrik~O Hoyer. 2002.
\newblock Non-negative sparse coding.
\newblock In \emph{Neural Networks for Signal Processing, 2002. Proceedings of
  the 2002 12th IEEE Workshop on}, pages 557--565. IEEE.

\bibitem[{Huth et~al.(2016)Huth, de~Heer, Griffiths, Theunissen, and
  Gallant}]{huth2016natural}
Alexander~G Huth, Wendy~A de~Heer, Thomas~L Griffiths, Fr{\'e}d{\'e}ric~E
  Theunissen, and Jack~L Gallant. 2016.
\newblock Natural speech reveals the semantic maps that tile human cerebral
  cortex.
\newblock \emph{Nature}, 532(7600):453--458.

\bibitem[{Karpathy and Li(2015)}]{karpathy2015deep}
Andrej Karpathy and Fei-Fei Li. 2015.
\newblock Deep visual-semantic alignments for generating image descriptions.
\newblock In \emph{Proceedings of the IEEE conference on computer vision and
  pattern recognition}, pages 3128--3137.

\bibitem[{Kiela and Bottou(2014)}]{kiela2014learning}
Douwe Kiela and L{\'e}on Bottou. 2014.
\newblock Learning image embeddings using convolutional neural networks for
  improved multi-modal semantics.
\newblock In \emph{Proceedings of the 2014 Conference on Empirical Methods in
  Natural Language Processing (EMNLP)}, pages 36--45.

\bibitem[{Kriegeskorte et~al.(2008)Kriegeskorte, Mur, and
  Bandettini}]{kriegeskorte2008representational}
Nikolaus Kriegeskorte, Marieke Mur, and Peter~A Bandettini. 2008.
\newblock Representational similarity analysis-connecting the branches of
  systems neuroscience.
\newblock \emph{Frontiers in systems neuroscience}, 2:4.

\bibitem[{Krizhevsky et~al.(2012)Krizhevsky, Sutskever, and
  Hinton}]{krizhevsky2012imagenet}
Alex Krizhevsky, Ilya Sutskever, and Geoffrey~E Hinton. 2012.
\newblock Imagenet classification with deep convolutional neural networks.
\newblock In \emph{Advances in neural information processing systems}, pages
  1097--1105.

\bibitem[{Lazaridou et~al.(2015)Lazaridou, Pham, and
  Baroni}]{lazaridou2015combining}
Angeliki Lazaridou, Nghia~The Pham, and Marco Baroni. 2015.
\newblock Combining language and vision with a multimodal skip-gram model.
\newblock \emph{arXiv preprint arXiv:1501.02598}.

\bibitem[{Lucy and Gauthier(2017)}]{lucy2017distributional}
Li~Lucy and Jon Gauthier. 2017.
\newblock Are distributional representations ready for the real world?
  evaluating word vectors for grounded perceptual meaning.
\newblock \emph{arXiv preprint arXiv:1705.11168}.

\bibitem[{Mairal et~al.(2010)Mairal, Bach, Ponce, and
  Sapiro}]{mairal2010online}
Julien Mairal, Francis Bach, Jean Ponce, and Guillermo Sapiro. 2010.
\newblock Online learning for matrix factorization and sparse coding.
\newblock \emph{Journal of Machine Learning Research}, 11(Jan):19--60.

\bibitem[{McRae et~al.(2005)McRae, Cree, Seidenberg, and
  McNorgan}]{mcrae2005semantic}
Ken McRae, George~S Cree, Mark~S Seidenberg, and Chris McNorgan. 2005.
\newblock Semantic feature production norms for a large set of living and
  nonliving things.
\newblock \emph{Behavior research methods}, 37(4):547--559.

\bibitem[{Mikolov et~al.(2013)Mikolov, Sutskever, Chen, Corrado, and
  Dean}]{mikolov2013distributed}
Tomas Mikolov, Ilya Sutskever, Kai Chen, Greg~S Corrado, and Jeff Dean. 2013.
\newblock Distributed representations of words and phrases and their
  compositionality.
\newblock In \emph{Advances in neural information processing systems}, pages
  3111--3119.

\bibitem[{Miller(1995)}]{miller1995wordnet}
George~A Miller. 1995.
\newblock Wordnet: a lexical database for english.
\newblock \emph{Communications of the ACM}, 38(11):39--41.

\bibitem[{Mitchell et~al.(2008)Mitchell, Shinkareva, Carlson, Chang, Malave,
  Mason, and Just}]{mitchell2008predicting}
Tom~M Mitchell, Svetlana~V Shinkareva, Andrew Carlson, Kai-Min Chang, Vicente~L
  Malave, Robert~A Mason, and Marcel~Adam Just. 2008.
\newblock Predicting human brain activity associated with the meanings of
  nouns.
\newblock \emph{science}, 320(5880):1191--1195.

\bibitem[{Murphy et~al.(2012)Murphy, Talukdar, and
  Mitchell}]{murphy2012learning}
Brian Murphy, Partha Talukdar, and Tom Mitchell. 2012.
\newblock Learning effective and interpretable semantic models using
  non-negative sparse embedding.
\newblock \emph{Proceedings of COLING 2012}, pages 1933--1950.

\bibitem[{Ngiam et~al.(2011)Ngiam, Khosla, Kim, Nam, Lee, and
  Ng}]{ngiam2011multimodal}
Jiquan Ngiam, Aditya Khosla, Mingyu Kim, Juhan Nam, Honglak Lee, and Andrew~Y
  Ng. 2011.
\newblock Multimodal deep learning.
\newblock In \emph{Proceedings of the 28th international conference on machine
  learning (ICML-11)}, pages 689--696.

\bibitem[{Oquab et~al.(2014)Oquab, Bottou, Laptev, and
  Sivic}]{oquab2014learning}
Maxime Oquab, Leon Bottou, Ivan Laptev, and Josef Sivic. 2014.
\newblock Learning and transferring mid-level image representations using
  convolutional neural networks.
\newblock In \emph{Computer Vision and Pattern Recognition (CVPR), 2014 IEEE
  Conference on}, pages 1717--1724. IEEE.

\bibitem[{Pennington et~al.(2014)Pennington, Socher, and
  Manning}]{pennington2014glove}
Jeffrey Pennington, Richard Socher, and Christopher Manning. 2014.
\newblock Glove: Global vectors for word representation.
\newblock In \emph{Proceedings of the 2014 conference on empirical methods in
  natural language processing (EMNLP)}, pages 1532--1543.

\bibitem[{Senel et~al.(2017)Senel, Utlu, Yucesoy, Koc, and
  Cukur}]{senel2017semantic}
Lutfi~Kerem Senel, Ihsan Utlu, Veysel Yucesoy, Aykut Koc, and Tolga Cukur.
  2017.
\newblock Semantic structure and interpretability of word embeddings.
\newblock \emph{arXiv preprint arXiv:1711.00331}.

\bibitem[{Silberer et~al.(2017)Silberer, Ferrari, and
  Lapata}]{silberer2017visually}
Carina Silberer, Vittorio Ferrari, and Mirella Lapata. 2017.
\newblock Visually grounded meaning representations.
\newblock \emph{IEEE transactions on pattern analysis and machine
  intelligence}, 39(11):2284--2297.

\bibitem[{S{\o}gaard(2016)}]{sogaard2016evaluating}
Anders S{\o}gaard. 2016.
\newblock Evaluating word embeddings with fmri and eye-tracking.
\newblock In \emph{Proceedings of the 1st Workshop on Evaluating Vector-Space
  Representations for NLP}, pages 116--121.

\bibitem[{Von~Ahn and Dabbish(2004)}]{von2004labeling}
Luis Von~Ahn and Laura Dabbish. 2004.
\newblock Labeling images with a computer game.
\newblock In \emph{Proceedings of the SIGCHI conference on Human factors in
  computing systems}, pages 319--326. ACM.

\bibitem[{Xu et~al.(2016)Xu, Murphy, and Fyshe}]{xu2016brainbench}
Haoyan Xu, Brian Murphy, and Alona Fyshe. 2016.
\newblock Brainbench: A brain-image test suite for distributional semantic
  models.
\newblock In \emph{Proceedings of the 2016 Conference on Empirical Methods in
  Natural Language Processing}, pages 2017--2021.

\bibitem[{Zeiler and Fergus(2013)}]{zeilerfergus2013}
Matthew~D. Zeiler and Rob Fergus. 2013.
\newblock Visualizing and understanding convolutional networks.
\newblock \emph{CoRR}, abs/1311.2901.

\end{thebibliography}

\end{document}